\pdfoutput=1
\documentclass[10pt,twocolumn,letterpaper]{article}

\usepackage[pagenumbers]{cvpr} 

\usepackage{graphicx}
\usepackage{amsmath}
\usepackage{amssymb}
\usepackage{booktabs}

\usepackage[pagebackref,breaklinks,colorlinks]{hyperref}

\usepackage{amsmath}
\usepackage{amsfonts}
\usepackage{stmaryrd}
\usepackage[table,xcdraw]{xcolor}
\usepackage{pifont}
\usepackage{multirow}
\usepackage{makecell}
\usepackage{endnotes}
\usepackage[switch]{lineno}

\usepackage{booktabs}
\usepackage{colortbl}
\definecolor{colorh}{rgb}{1,0.60,0.20}
\definecolor{colorm}{rgb}{1,0.72,0.30}
\definecolor{colorl}{rgb}{1,0.88,0.70}
\newcommand{\colorh}[1]{\colorbox{colorh}{{#1}}}
\newcommand{\colorm}[1]{\colorbox{colorm}{{#1}}}
\newcommand{\colorl}[1]{\colorbox{colorl}{{#1}}}

\usepackage[capitalize]{cleveref}
\crefname{section}{Sec.}{Secs.}
\Crefname{section}{Section}{Sections}
\Crefname{table}{Table}{Tables}
\crefname{table}{Tab.}{Tabs.}

\def\cvprPaperID{*****} 
\def\confName{CVPR}
\def\confYear{2023}

\begin{document}

\title{3D-STMN: Dependency-Driven Superpoint-Text Matching Network for End-to-End 3D Referring Expression Segmentation}
\author{
    Changli Wu$^{1}$,
    Yiwei Ma$^{1}$,
    Qi Chen$^{1}$,
    Haowei Wang$^{1}$,
    Gen Luo$^{1}$,
    Jiayi Ji$^{1}$\thanks{Corresponding author.},
    Xiaoshuai Sun$^{1, 2}$ \\
    $^1$Key Laboratory of Multimedia Trusted Perception and Efficient Computing, Ministry of Education of China, \\
    School of Informatics, Xiamen University. 
    $^2$Institute of Artificial Intelligence, Xiamen University. \\
    {\tt\small \{wuchangli, yiweima, chenqi, wanghaowei, luogen\}@stu.xmu.edu.cn, jjyxmu@gmail.com, xssun@xmu.edu.cn}
}
\maketitle

\begin{abstract}
In 3D Referring Expression Segmentation (3D-RES), the earlier approach adopts a two-stage paradigm, extracting segmentation proposals and then matching them with referring expressions. However, this conventional paradigm encounters significant challenges, most notably in terms of the generation of lackluster initial proposals and a pronounced deceleration in inference speed.
Recognizing these limitations, we introduce an innovative end-to-end Superpoint-Text Matching Network (3D-STMN) that is enriched by dependency-driven insights. One of the keystones of our model is the Superpoint-Text Matching (STM) mechanism. Unlike traditional methods that navigate through instance proposals, STM directly correlates linguistic indications with their respective superpoints, clusters of semantically related points. This architectural decision empowers our model to efficiently harness cross-modal semantic relationships, primarily leveraging densely annotated superpoint-text pairs, as opposed to the more sparse instance-text pairs.
In pursuit of enhancing the role of text in guiding the segmentation process, we further incorporate the Dependency-Driven Interaction (DDI) module to deepen the network's semantic comprehension of referring expressions. Using the dependency trees as a beacon, this module discerns the intricate relationships between primary terms and their associated descriptors in expressions, thereby elevating both the localization and segmentation capacities of our model.
Comprehensive experiments on the ScanRefer benchmark reveal that our model not only set new performance standards, registering an mIoU gain of 11.7 points but also achieve a staggering enhancement in inference speed, surpassing traditional methods by 95.7 times. The code and models are available at \url{https://github.com/sosppxo/3D-STMN}.
\end{abstract}

\begin{figure}[!t]
\centering 
  \includegraphics[width=1.0\columnwidth]{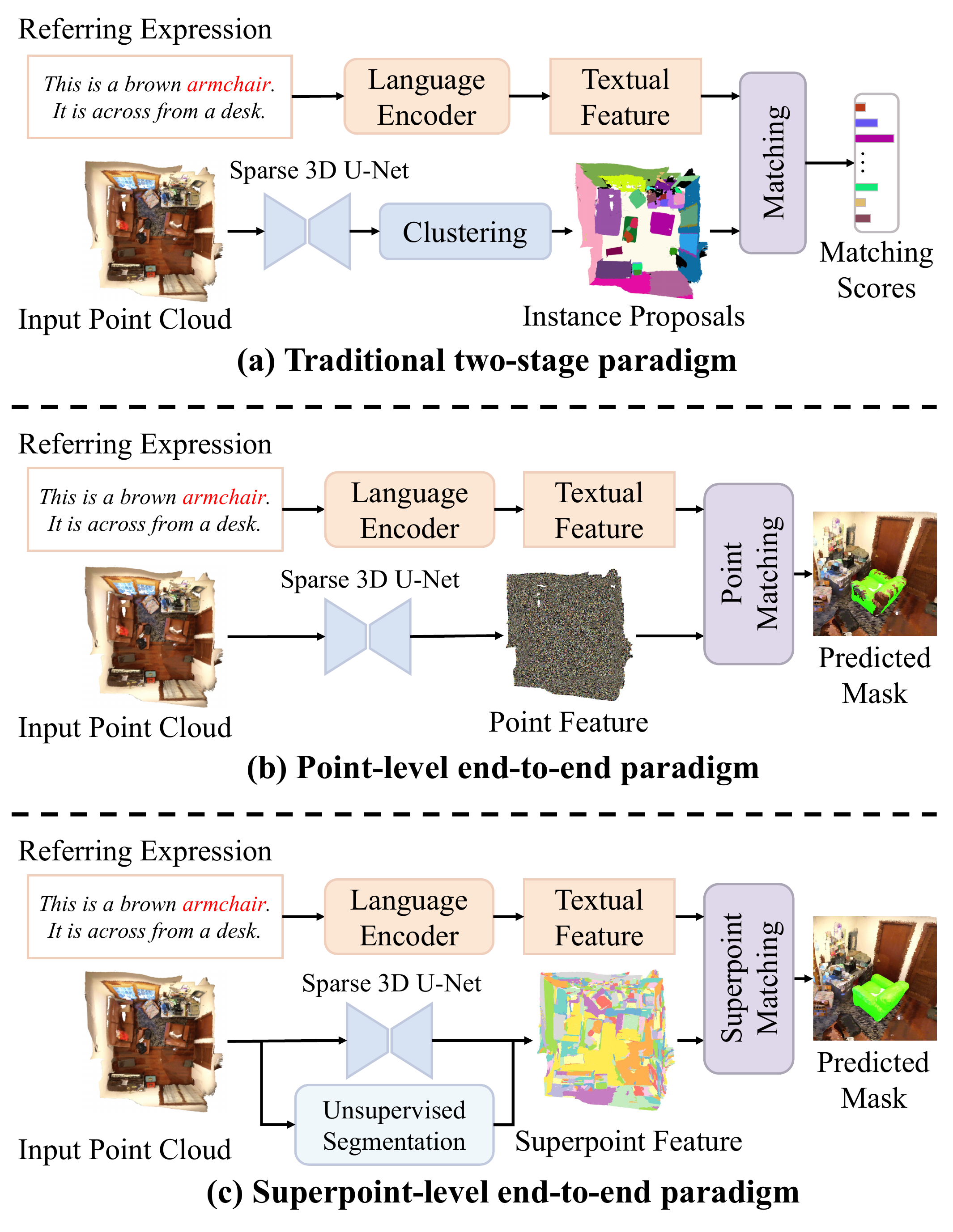}
  \caption{ A comparison among (a) traditional two-stage paradigm, (b) point-level end-to-end paradigm, and (c) superpoint-level end-to-end paradigm.
  }
  \label{fig: fig1}
\end{figure}

\section{Introduction}
\label{sec:intro}
The goal of 3D visual grounding is to locate instances within a 3D scene based on given natural language descriptions~\cite{Chen_Chang_Nießner_2020}. In recent years, it has become a hot topic in academic research due to its wide-ranging application scenarios, including autonomous robotics, human-machine interaction, and self-driving systems, among others. Within this field, the task of 3D Referring Expression Segmentation (3D-RES) emerges as a formidable challenge. Compared to 3D visual detection tasks~\cite{Wang_Ye_Cao_Huang_Sun_He_Tao, He_Li_Li_Zhang, Chen_Chang_Nießner_2020, Achlioptas_Abdelreheem_Xia_Elhoseiny_Guibas_2020, Zhao_Cai_Sheng_Xu_2021, Luo_Fu_Kong_Gao_Ren_Shen_Xia_Liu_2022}, which merely locate the target objects with bounding boxes, 3D-RES demands a more complex understanding. It not only requires the identification of target instances within sparse point clouds, but it also requires the provision of precise 3D masks that correspond to each identified target instance.

At present, the only existing method referred to as TGNN~\cite{Huang_Lee_Chen_Liu_2021}, operates in a two-stage manner. In the initial stage, an independent text-agnostic segmentation model is trained to generate instance proposals. Then, in the second stage, a graph neural network is employed to forge links between the generated proposals and textual descriptions, as shown in Fig.~\ref{fig: fig1}-(a). Despite achieving good results, this two-stage paradigm still suffers from three primary issues:
(1) The decoupling of segmentation from matching creates an over-reliance on the preliminary text-independent segmentation outcomes. Any inaccuracies or omissions in the first phase can insurmountably compromise the accuracy of the subsequent matching phase, irrespective of its intrinsic efficacy.
(2) The model overlooks the inherent hierarchical and dependency structures within the referring sentence. Its linear language modeling strategy falls short in capturing intricate semantic nuances, leading to missteps in both localization and segmentation.
(3) To amplify the recall efficiency in the secondary phase, the first stage extracts dense candidate masks through iterative clustering over several stages. This iterative process considerably decelerates the model's inference speed. Hence, despite its merits, the two-stage paradigm employed by TGNN leaves room for substantial improvement in both accuracy and efficiency.

A natural approach would be to employ an end-to-end method that directly matches textual features with points in the 3D point cloud, as shown in Fig.~\ref{fig: fig1}-(b). This approach has been widely proven effective in 2D-RES tasks ~\cite{Ye_Rochan_Liu_Wang_2019, Liu_Zhang_Zha_Wu_2019, Ding_Liu_Wang_Jiang_2021, yang2022lavt}. However, it has not been translated well to sparse, irregular 3D point cloud data, as it results in a low recall rate.
As a solution, 3D-SPS~\cite{Luo_Fu_Kong_Gao_Ren_Shen_Xia_Liu_2022} in 3D visual detection has suggested a method that progressively selects keypoints guided by language and regresses boxes using this keypoint information. However, this approach disrupts the continuity of the 3D mask in 3D-RES tasks, thereby deteriorating the segmentation results.

To tackle the aforementioned challenges, we present a dependency-driven Superpoint-Text Matching Network for an end-to-end 3D-RES. The idea of our approach is the matching of the expressions with over-segmented superpoints~\cite{Landrieu_Simonovsky_2018}. As illustrated in Fig.~\ref{fig: fig1}-(c), these superpoints are initially aggregated through a clustering algorithm, thus attaining fine-grained semantic units. These superpoints, embodying semantics and being significantly fewer compared to the unordered points in a 3D point cloud, offer advantages in performance and speed during the matching process. In contrast to the proposals in TGNN, superpoints are fine-grained units derived from over-segmentation, capable of covering the entire scene, thereby averting the issues of inaccurate segmentation or missing instances. In light of this, we introduce a new Superpoint-Text Matching (STM) mechanism for 3D-RES, leveraging the aggregation of text features from superpoints to acquire the mask of the target instance. To bolster semantic parsing from the textual perspective, we devise a Dependency-Driven Interaction (DDI) module, achieving token-level interactions. This module exploits the prior information from the dependency syntax tree to steer the flow of text information. This structure further enhances inference on relationships among different instances via the network architecture, thus markedly improving the model's segmentation ability. We have conducted extensive quantitative and qualitative experiments on the classic ScanRefer dataset for investigation. It's noteworthy that our method achieves a remarkable 95.7-fold increase in inference speed while outperforming TGNN by an impressive 11.7 points.

To summarize, our main contributions are as follows:
\begin{itemize}
    \item We propose a novel efficient end-to-end framework 3D-STMN based on Superpoint-Text Matching (STM) mechanism for aligning superpoint with textual modality, making superpoint a highly competitive player in multimodal representation.
    \item We design a Dependency-Driven Interaction (DDI) module to exploit the prior information from the dependency syntax tree to steer the flow of text information, markedly improving the model's segmentation ability.
    \item Extensive experiments show that our method significantly outperforms the previous two-stage baseline in the ScanRefer benchmark, registering a mIoU gain of 11.7 points but also achieving a staggering enhancement in inference speed.
\end{itemize}

\begin{figure*}[!t]
\centering 
    \includegraphics[width=2\columnwidth]{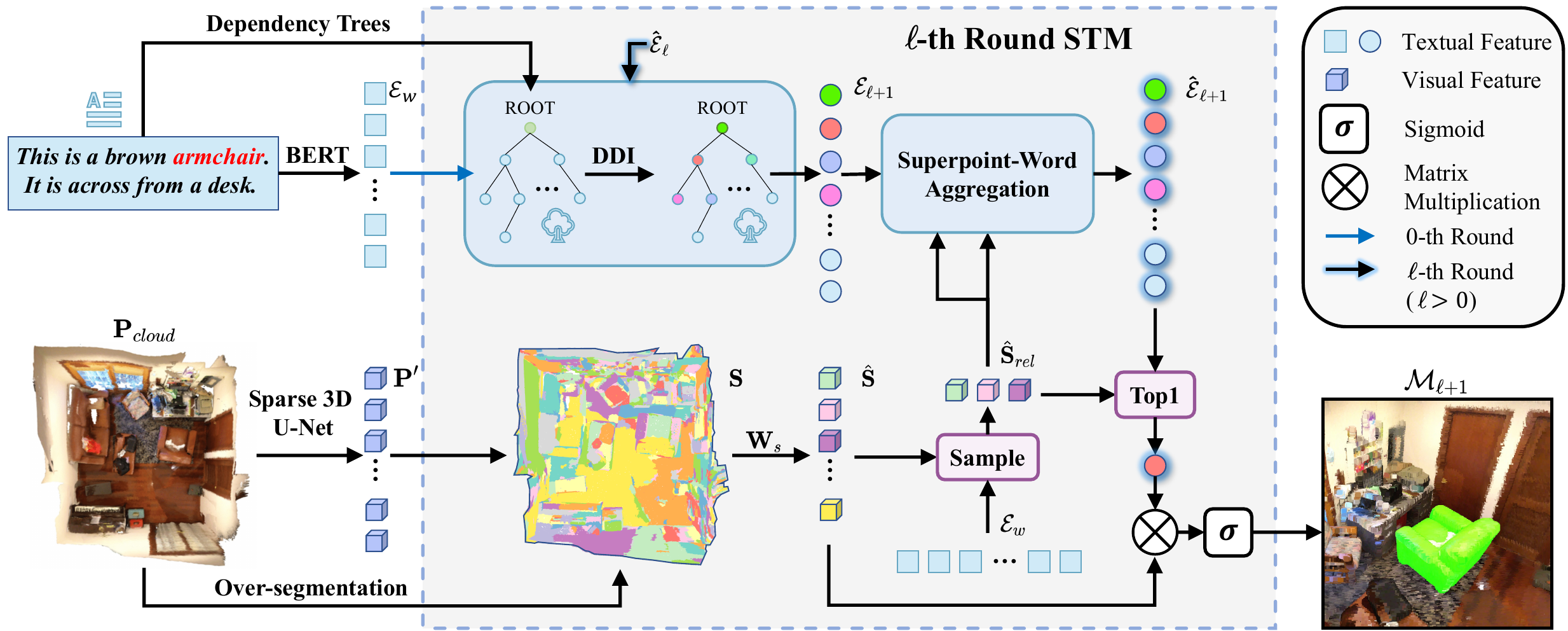}
  \caption{ Overview of our 3D Superpoint-Text Matching Network (3D-STMN).
  }
  \label{fig: fig2}
\end{figure*}

\section{Related Work}
\subsection{2D Referring Expression Comprehension and Segmentation}
2D-REC tasks involve predicting a bounding box that corresponds to the object described in a given referring expression~\cite{Nagaraja_Morariu_Davis_2016, Yu_Poirson_Yang_Berg_Berg_2016, Hu_Rohrbach_Andreas_Darrell_Saenko_2017, Yu_Tan_Bansal_Berg_2017, Deng_Wu_Wu_Hu_Lyu_Tan_2018, Zhuang_Wu_Shen_Reid_Hengel_2017, Sadhu_Chen_Nevatia_2019, Yang_Li_Yu_2020, luo2020multi}. In contrast, 2D-RES aims to predict a segmentation mask that accurately delineates the referred object, in order to achieve more precise localization results~\cite{Hu_Rohrbach_Darrell_2016, Yu_Lin_Shen_Yang_Lu_Bansal_Berg_2018, Ye_Rochan_Liu_Wang_2019, Shi_Li_Meng_Wu_2018}.
Many of these works adopt a two-stage paradigm of segmentation followed by matching~\cite{Li_Li_Kuo_Shu_Qi_Shen_Jia_2018, Margffoy-Tuay_Pérez_Botero_Arbeláez_2018}. While these methods' performance is ultimately limited by the quality of the segmentation models, a number of approaches have been developed that refine segmentation masks using a single-stage network~\cite{Ye_Rochan_Liu_Wang_2019, Liu_Zhang_Zha_Wu_2019, luo2020cascade}.
Although these works have shown promising results in 2D-REC and 2D-RES tasks, they cannot be directly applied to 3D point cloud scenes due to the inherent challenges posed by the sparse and irregular, nature of 3D point clouds.

\subsection{3D Referring Expression Comprehension and Segmentation}
Recently, 3D REC has garnered significant attention, aiming to localize objects within a 3D scene based on referring expressions. ScanRefer~\cite{Chen_Chang_Nießner_2020} provides a dataset based on ScanNetv2~\cite{Dai_Chang_Savva_Halber_Funkhouser_Niessner_2017} for the Referring 3D Instance Localization task. Additionally, ReferIt3D~\cite{Achlioptas_Abdelreheem_Xia_Elhoseiny_Guibas_2020} proposes two datasets, Sr3D and Nr3D. Most existing methods~\cite{Chen_Chang_Nießner_2020, Achlioptas_Abdelreheem_Xia_Elhoseiny_Guibas_2020, Zhao_Cai_Sheng_Xu_2021, Yuan_Yan_Liao_Zhang_Wang_Li_Cui_2021, Yang_Zhang_Wang_Luo_2021, Huang_Lee_Chen_Liu_2021, Feng_Li_Li_Zhang_Zhang_Zhu_Zhang_Wang_Mian_2021} adopt a two-stage paradigm. Meanwhile, some methods have adopted a framework supporting single-stage training~\cite{Luo_Fu_Kong_Gao_Ren_Shen_Xia_Liu_2022, Jain_Gkanatsios_Mediratta_Fragkiadaki_2021, Wu_Cheng_Zhang_Cheng_Zhang_2022}.
To address 3D-RES tasks, TGNN~\cite{Huang_Lee_Chen_Liu_2021} proposed a two-stage model. 
However, both the accuracy and inference speed of TGNN exhibit inherent limitations of the segmentation model. To circumvent these challenges, we propose an end-to-end Superpoint-Text Matching Network in this paper.

\subsection{Superpoint based 3D Scene Understanding}

Similar to superpixels~\cite{Achanta_Shaji_Smith_Lucchi_Fua_Süsstrunk_2012, Tu_Liu_Jampani_Sun_Chien_Yang_Kautz_2018}, superpoints have been used for 3D point cloud segmentation ~\cite{Papon_Abramov_Schoeler_Worgotter_2013, Lin_Wang_Zhai_Li_Li_2018, Landrieu_Simonovsky_2018, Robert_Raguet_Landrieu_2023} and object detection~\cite{Han_Zheng_Xu_Fang_2020, Engelmann_Bokeloh_Fathi_Leibe_NieBner_2020}. For 3D instance segmentation, superpoints have also demonstrated incredible potential~\cite{Liang_Li_Xu_Tan_Jia_2021, Sun_Qing_Tan_Xu_2022}.
However, these works only applied superpoint to pure visual tasks and did not explore the ability of superpoint to align with textual modality. In this paper, we first propose a framework for aligning superpoint with textual modality, making superpoint a highly competitive player in multimodal representations.

\section{Method} \label{method}
In this section, we provide a comprehensive overview of the 3D-STMN. The framework is illustrated in Fig.~\ref{fig: fig2}.
First, the features of visual and linguistic modalities are extracted in parallel (Sec.~\ref{sec: feat}). 
Next, we demonstrate how the 3D-RES task can be formulated as a Superpoint-Text matching problem (Sec.~\ref{sec: match}).
Concurrently, we detail the proposed Dependency-Driven Interaction module (Sec.~\ref{sec: ddi}). 
Finally, we outline the training objectives of 3D-SPMN (Sec.~\ref{sec: loss}).

\subsection{Feature Extraction} \label{sec: feat}
\subsubsection{Visual Modality}
Given a point cloud scene with $\mathcal{N}_p$ points, it can be represented as 
$\mathbf{P}_{cloud}\in \mathbb{R}^{\mathcal{N}_p\times (3+F)}$. 
Here, each point comes with 3D coordinates along with an \textit{F}-dimensional auxiliary feature that includes RGB, normal vectors, among others. Building on TGNN~\cite{Huang_Lee_Chen_Liu_2021}, we employ a singular Sparse 3D U-Net~\cite{Graham_Engelcke_Maaten_2018} to extract point-wise features, represented as $\mathbf{P'}\in \mathbb{R}^{\mathcal{N}_p\times C_p}$.

\subsubsection{Linguistic Modality}

Given a free-form plain text description of the target object with $\mathcal{N}_w$ words $\{c_i\}_{i=1}^{\mathcal{N}_w}$, we follow~\cite{Huang_Lee_Chen_Liu_2021}  to adopt a pre-trained BERT~\cite{Devlin_Chang_Lee_Toutanova_2019} to extract the $C_t$-dimensional word-level embeddings $\mathbf{\mathcal{E}}_w \in \mathbb{R}^{\mathcal{N}_w\times C_t}$, and description-level embedding $\mathbf{d}_{0} \in \mathbb{R}^{C_t}$ which is the embeddings of [CLS] token.

\subsection{Superpoint-Text Matching} \label{sec: match}

\subsubsection{Superpoints and Dependency-Driven Text}
After extracting the features, we perform over-segmentation to $\mathbf{P}_{cloud}$ to obtain $\mathcal{N}_s$ superpoints $\{\mathcal{K}_i\}^{\mathcal{N}_s}_{i=1}$\cite{Landrieu_Simonovsky_2018}.

To obtain the superpoint-level features $\mathbf{S} \in \mathbb{R}^{\mathcal{N}_s\times C_p}$, we directly feed point-wise features $\mathbf{P'}$ into superpoint pooling layer based on $\{\mathcal{K}_i\}^{\mathcal{N}_s}_{i=1}$, which can be formulated as:
\begin{equation}
    \mathbf{S}^i = \text{AvgPool}(\mathbf{P'},\mathcal{K}_i),
\end{equation}
where $\mathbf{S}^i$ denotes the feature of the $i$-th superpoint, $\mathcal{K}_i$ denotes the set of indices of points contained in the $i$-th superpoint, $\text{AvgPool}(\cdot)$ denotes the superpoint average pooling operation.

To the text end, we feed the expression with word-level embeddings $\mathbf{\mathcal{E}}_w$ into the proposed DDI module which aims to construct a description-dependency graph 
and outputs the dependency-driven feature $\mathbf{\mathcal{E}}_0$, which is formulated as follows:
\begin{eqnarray}
    \mathbf{\hat{\mathcal{E}}}_1 &=& (\mathbf{\mathcal{E}}_{\text{root}}\ \|\ \mathbf{\mathcal{E}}_w)\mathbf{W}_t, \\
    \mathbf{\mathcal{E}}_1 &=& \text{DDI}(\mathbf{\hat{\mathcal{E}}}_1),
\end{eqnarray}
where $\mathbf{W}_t\in\mathbb{R}^{C_t\times D}$ is a learnable parameter, $\mathbf{\mathcal{E}}_w\in \mathbb{R}^{\mathcal{N}_w\times C_t}$, $\mathbf{\mathcal{E}}_1\in \mathbb{R}^{(\mathcal{N}_w+1)\times D}$, $\mathbf{\mathcal{E}}_{\text{root}}$ denotes the randomly initialized ROOT node feature, $\|$ denotes  the concatenation operation. More details about the DDI module are presented in Sec.~\ref{sec: ddi}.

To enhance the efficiency of subsequent processing, we adopt a filtering approach on $\mathbf{S}$ after linear projection, which is widely used in multimodal segmentation tasks~\cite{Ding_Ding_Hui_Huang_Wei_Wei_Liu_2022, Luo_Fu_Kong_Gao_Ren_Shen_Xia_Liu_2022}. Specifically, we acquire the $k_{rel}$ superpoints based on the relevance score $s_r$ between the superpoints and their corresponding descriptions.  The filtering process can be given by: 
\begin{eqnarray}
    \hat{\mathbf{S}} &=& \mathbf{S}\mathbf{W}_{s}, \\
    \mathbf{\mathcal{A}}&=& \text{softmax}\Big( \frac{\hat{\mathbf{S}}\mathbf{Q}_s \cdot (\mathbf{\mathcal{E}}_w\mathbf{K}_t)^T}{\sqrt{D}} \Big),\\
    s_r^i &=& \sum_{j=1}^{\mathcal{N}_w}\mathbf{\mathcal{A}}^{ij},\\
    \hat{\mathbf{S}}_{rel} &=& \hat{\mathbf{S}}[\text{ArgTopk}(s_r,k_{rel})]\ \| \ \text{AvgPool}(\hat{\mathbf{S}}),
\end{eqnarray}
where $\mathbf{W}_{s}\in \mathbb{R}^{C_p\times D}, \mathbf{Q}_s\in \mathbb{R}^{D\times D}, \mathbf{K}_t \in \mathbb{R}^{C_t\times D}$ denote learnable parameters.

$\text{AvgPool}(\hat{\mathbf{S}})$ plays the role of golobal features, and $\|$ denotes concatenation. $\hat{\mathbf{S}}_{rel}\in \mathbb{R}^{(k_{rel}+1)\times D}$ denotes features of description-relevant superpoints.

\subsubsection{Superpoint-Text Matching Process} \label{sec: STM}
To perform Superpoint-Text matching, we initially project the superpoint features $\mathbf{S}$ to a $D$-dimensional subspace that corresponds to the text embedding $\mathbf{\mathcal{E}}$. After a description-guided sampling of superpoints, we update the embedding of each text token using Superpoints-Word Aggregation (SWA) with adaptive attention weights. 
We design it as a multi-round refinement process, which can be described as follows:

\begin{eqnarray}
    \mathbf{\mathcal{E}}_{\ell+1} &=& \text{DDI}(\hat{\mathbf{\mathcal{E}}}_{\ell}),\\
    \hat{\mathbf{\mathcal{E}}}_{\ell+1} &=& \text{SWA}(\mathbf{\mathcal{E}}_{\ell+1}, \hat{\mathbf{S}}_{rel}), \ell=0,1,...,L-1,
\end{eqnarray}
where $\hat{\mathbf{\mathcal{E}}}_{\ell}, \hat{\mathbf{\mathcal{E}}}_{\ell+1}\in \mathbb{R}^{(\mathcal{N}_w+1)\times D}$ and $L$ is the number of multiple rounds. The details about SWA are presented in the following subsection.

Next, we perform matrix multiplication between $\hat{\mathbf{S}}$ and $\hat{\mathbf{\mathcal{E}}}$ to obtain the response maps that capture the relationship between all superpoints and word tokens. This computation can be described as follows:
\begin{equation}
    \mathbf{M}_{\ell+1} = \sigma(\hat{\mathbf{\mathcal{E}}}_{\ell+1}\hat{\mathbf{S}}^T),
\end{equation}
where $\hat{\mathbf{S}}^T \in \mathbb{R}^{D\times \mathcal{N}_s}$ is the transpose of $\hat{\mathbf{S}}$, $\mathbf{M}_{\ell+1}\in \mathbb{R}^{(\mathcal{N}_w+1)\times \mathcal{N}_s}$ is the response maps, and $\sigma(\cdot)$ denotes sigmoid function. In particular, $\mathbf{M}_{\ell+1}^n \in \mathbb{R}^{\mathcal{N}_s}$ is the response map of the $n$-th token, based on which we can generate the segmentation result and attention mask $ \mathbf{A}_{\ell+1}^n\in \mathbb{R}^{\mathcal{N}_s}$ corresponding to the $n$-th token. 

To obtain the final mask, we choose the response map $\mathbf{\mathcal{M}}_{\ell+1} \in \mathbb{R}^{\mathcal{N}_s}$ associated with the word token that has the highest correlation score with all description-relevant superpoints:
\begin{eqnarray}
    \mathbf{A}_{v,\ell+1}&=& \text{softmax}\Big(\frac{\hat{\mathbf{\mathcal{E}}}_{\ell+1}\mathbf{Q}_t^{\ell+1} \cdot (\hat{\mathbf{S}}_{rel}\mathbf{K}_s^{\ell+1})^T}{\sqrt{D}}\Big),\\
    \mathbf{s}_v^i &=& \sum_{j=1}^{k_{rel}+1}\mathbf{A}_{v, \ell+1}^{ij},\\
    \mathbf{\mathcal{M}}_{\ell+1} &=& \mathbf{M}_{\ell+1}[\text{ArgMax}(\mathbf{s}_v)],
\end{eqnarray}
where $\text{ArgMax}(\cdot)$ returns the index corresponding to the maximum value. $\mathbf{Q}_t^{\ell+1}, \mathbf{K}_s^{\ell+1}\in \mathbb{R}^{D\times D}$ are learnable parameters. $\mathbf{A}_{v,\ell}^{ij}$ means the attention score between the $i$-th word and the $j$-th description-relevant superpoint, and $\mathbf{s}_v^i$ denotes the visual correlation score of the $i$-th word. 

\subsubsection{Superpoint-Word Aggregation}\label{sec: swa}
To enhance the discriminative power of the textual segmentation kernel, we introduce a Superpoint-Word Aggregation module, which is designed to refine the multi-round modality interaction between superpoints and textual descriptions.

At the $\ell$-th layer, SWA adaptively aggregates the superpoint features to enables each word to absorb the visual information of the related superpoint features. 

As depicted in Fig.~\ref{fig: fig2}, the adaptive superpoint-word cross-attention block utilizes the dependency-driven feature $\mathbf{\mathcal{E}}$ to refine the word features by incorporating information from the related superpoints: 
\begin{small}
\begin{equation}
    \hat{\mathbf{\mathcal{E}}}_{\ell+1} = \text{softmax}\Big(\frac{\mathbf{\mathcal{E}}_{\ell+1}\mathbf{Q}_{\ell+1}\cdot (\hat{\mathbf{S}}_{rel}\mathbf{K}_{\ell+1})^T}{\sqrt{D}}+\mathbf{A}_{\ell}\Big)\cdot \hat{\mathbf{S}}_{rel}\mathbf{V}_{\ell+1},
\end{equation}
\end{small}
where $\hat{\mathbf{\mathcal{E}}}_{\ell+1}\in \mathbb{R}^{(\mathcal{N}_w+1)\times D}$is the output of superpoint-word cross-attention. $\mathbf{Q}_{\ell+1}, \mathbf{K}_{\ell+1}, \mathbf{V}_{\ell+1} \in \mathbb{R}^{D \times D}$ are learnable parameters. 

$\mathbf{A}_{\ell}\in \mathbb{R}^{(\mathcal{N}_w+1)\times(k_{rel}+1)}$ is superpoint attention masks. Given the predicted superpoint masks $\mathbf{M}_{\ell}$ from the prediction head, superpoint attention masks $\mathbf{A}_{\ell}$ filter superpoint with a threshold $\tau$, as
\begin{equation}
    \mathbf{A}_{\ell}^{ij}=\left\{
    \begin{aligned}
    0   &\quad \text{if}\quad \mathbf{M}_{\ell}^{ij}\geq \tau \\
    -\infty   &\quad \text{otherwise}
    \end{aligned}
    \right.
    .
\end{equation}
$\mathbf{A}_{\ell}^{ij}$ indicates $i$-th word token attending to $j$-th superpoint where $\mathbf{M}_{\ell}^{ij}$ is higher than $\tau$. Empirically, we set $\tau$ to 0.5. With transformer decoder layer stacking, superpoint attention masks $\mathbf{A}_{\ell}$ adaptively constrain cross-attention within the target instance.

\subsection{Dependency-Driven Interaction}\label{sec: ddi}

To explicitly decouple the textual description and effectively capture the dependency between words, we propose the Dependency-Driven Interaction module. 

\subsubsection{Description-Dependency Graph} \label{sec: DDG}

Given a free-form plain text description of the target object consisting of $\mathcal{N}_t$ sentences and a total of $\mathcal{N}_w$ words, we first use the Stanford CoreNLP~\cite{Manning_Surdeanu_Bauer_Finkel_Bethard_McClosky_2014} toolkit to obtain $\mathcal{N}_t$ dependency trees. Then we merge these $\mathcal{N}_t$ dependency trees into one graph by combining their ROOT nodes, as shown in Fig~\ref{fig: fig2}. Thus, for every description, the dependency graph has $\mathcal{N}_w+1$ nodes $\{u\}$ with $\mathcal{N}_w$ edges $\{e\}$. Each node represents a word including the special token ``ROOT'', while each edge represents a type of dependency relationship.

\subsubsection{Graph Transformer Layer with edge features}

Inspired by~\cite{Dwivedi_Bresson_2020}, we adopt a Graph Transformer Layer with edge features to more effectively leverage the abundant feature information available in Description-Dependency Graphs, which is stored in the form of edge attributes including dependency relationship. 
Given the textual features $\mathbf{\hat{\mathcal{E}}}_{0} = \{\mathbf{\hat{\mathcal{E}}}_{0}^0, \mathbf{\hat{\mathcal{E}}}_{0}^1, \cdots, \mathbf{\hat{\mathcal{E}}}_{0}^{\mathcal{N}_w+1}\}$, we directly derive the node features $\hat{\mathbf{h}}_i^0 = \{\hat{\mathbf{h}}_0^0, \hat{\mathbf{h}}_1^0, \cdots, \hat{\mathbf{h}}_{\mathcal{N}_w+1}^0\}$ based on their corresponding indices. For the edge features$\{\beta_{ij}\}$, we assign a unique ID to each dependency relationship which is passed via a linear projection to obtain $D$-dimensional hidden features $\mathbf{e}_{ij}^0$. 
\begin{eqnarray}
    \hat{\mathbf{h}}_i^{0} &=& \mathbf{\hat{\mathcal{E}}}_{0}^i, \\
    \mathbf{e}_{ij}^{0} &=& \mathbf{\beta}_{ij}\mathbf{B}^{0} + \mathbf{b}^{0} ,
\end{eqnarray}
where $\mathbf{B}^{0} \in \mathbb{R}^{1 \times D}$ and $\mathbf{b}^{0}\in \mathbb{R}^{D}$ are the parameters of the linear projection layers.

We now embed the pre-computed $k$-dimensional node positional encodings via a linear projection and add to the node features $\hat{\mathbf{h}}_i^{0}$.
\begin{equation}
\label{eqn:pe_embd_add}
{\lambda}_i^{0} =\mathbf{C}^{0} \lambda_i + c^{0} \;\  ; \;\ \mathbf{h}_i^{0} = \hat{\mathbf{h}}_i^{0} + {\lambda}_i^{0},  
\end{equation}

where $\mathbf{C}^{0} \in \mathbb{R}^{D \times k}$ and $c^{0}\in \mathbb{R}^{D}$
. Note that the Laplacian positional encodings are only added to the node features at the input layer and not during intermediate Graph Transformer layers.

Next, we proceed to define the update equations for the $\ell$-th layer. 

\begin{eqnarray}
    \mathbf{h}_i^{\ell} &=& \mathbf{\hat{\mathcal{E}}}_{\ell}^i, \\
    \mathbf{w}_{ij}^{\ell} &=& \textnormal{softmax}_j ( \hat{\mathbf{w}}_{ij}^{\ell} ), \label{eqn:softmax_edge_1}\\
    \hat{\mathbf{w}}_{ij}^{\ell} &=& \Big( \frac{ \mathbf{h}_i^{\ell}\mathbf{Q}_h^{\ell} \ \cdot \ \mathbf{h}_j^{\ell}\mathbf{K}_h^{\ell}}{\sqrt{D}} \Big)  \ \cdot \ e^{\ell}_{ij}\mathbf{E}_e^{\ell} , \label{eqn:softmax_edge_2}\\
    \hat{\mathbf{h}}_{i}^{\ell+1} &=& \Big(\sum_{j \in \mathcal{N}_i} \mathbf{w}_{ij}^{\ell} (\mathbf{h}_j^{\ell}\mathbf{V}_h^{\ell}) \Big)\mathbf{O}_h^{\ell} , \label{eqn:gt_layer_edge_h}\\
    \hat{\mathbf{e}}_{ij}^{\ell+1} &=& \hat{\mathbf{w}}_{ij}^{\ell}\mathbf{O}_e^{\ell}, \label{eqn:gt_layer_edge_e}
\end{eqnarray}
where $\mathbf{Q}_h^{\ell}, \mathbf{K}_h^{\ell}, \mathbf{V}_h^{\ell}, \mathbf{E}_e^{\ell}, \mathbf{O}_h^{\ell}, \mathbf{O}_e^{\ell} \in \mathbb{R}^{D \times D}$ denote learnable parameters. 

Considering the lack of long-range connections in dependency graph structures, we introduce self-attention mechanism and combine it with graph attention in parallel. The outputs $\hat{\mathbf{h}}_{i}^{\ell+1}$ are added by a self-attention outputs of $\mathbf{h}_i^\ell$ and succeeded by residual connections and normalization layers to get the outputs $\widetilde{\mathbf{h}}_i^{\ell+1}$. $\widetilde{\mathbf{h}}_i^{\ell+1}$ and $\hat{\mathbf{e}}_{ij}^{\ell+1}$ are then passed to separate Feed Forward Networks preceded and succeeded by residual connections and normalization layers, as:

\begin{eqnarray}
    \widetilde{\mathbf{h}}_{i}^{\ell+1} &=& \textnormal{Norm} \Big( \mathbf{h}_{i}^{\ell} + \textnormal{SA}(\mathbf{h}_i^\ell) + \hat{\mathbf{h}}_{i}^{\ell+1} \Big), \label{eqn:rc_norm1_h}\\
    \overline{\mathbf{h}}_{i}^{\ell+1} &=&  \textnormal{GeLU}( \widetilde{\mathbf{h}}_{i}^{\ell+1}\mathbf{W}_{h1}^{\ell})\mathbf{W}_{h2}^{\ell}, \label{eqn:ffn_h}\\
    \mathbf{h}_{i}^{\ell+1} &=& \textnormal{Norm} \Big( \widetilde{\mathbf{h}}_{i}^{\ell+1} + \overline{\mathbf{h}}_{i}^{\ell+1} \Big), \label{eqn:rc_norm2_h}\\
    \widetilde{\mathbf{e}}_{ij}^{\ell+1} &=& \textnormal{Norm} \Big( \mathbf{e}_{ij}^{\ell} + \hat{\mathbf{e}}_{ij}^{\ell+1} \Big), \label{eqn:rc_norm1_e}\\
    \overline{\mathbf{e}}_{ij}^{\ell+1} &=&  \textnormal{ReLU}(\widetilde{\mathbf{e}}_{ij}^{\ell+1}\mathbf{W}_{e1}^{\ell})\mathbf{W}_{e2}^{\ell}, \label{eqn:ffn_e}\\
    \mathbf{e}_{ij}^{\ell+1} &=& \textnormal{Norm} \Big( \widetilde{\mathbf{e}}_{ij}^{\ell+1} + \overline{\mathbf{e}}_{ij}^{\ell+1} \Big), \label{eqn:rc_norm2_e}
\end{eqnarray}
where $\mathbf{W}_{h1}^{\ell} \in \mathbb{R}^{D \times D_h}$, $\mathbf{W}_{h2}^{\ell} \in \mathbb{R}^{D_h \times D}$, $\mathbf{W}_{e1}^{\ell} \in \mathbb{R}^{D \times 2D}$, $\mathbf{W}_{e2}^{\ell} \in \mathbb{R}^{2D \times D}$ are learnable parameters,  $\widetilde{\mathbf{h}}_{i}^{\ell+1}, \overline{\mathbf{h}}_{i}^{\ell+1}$, $\widetilde{\mathbf{e}}_{ij}^{\ell+1}$, $\overline{\mathbf{e}}_{ij}^{\ell+1}$ denote intermediate representations,
$\textnormal{SA}(\mathbf{h}_i^\ell)$ means the $i$-th outputs of self-attention of $\hat{\mathbf{\mathcal{E}}}^{\ell}$.

Finally, the textual output of $\ell$-th layer DDI is obtained by concatenation of $\{\mathbf{h}_{i}^{\ell+1}\}_{i=1}^{\mathcal{N}_w + 1}$.
\begin{eqnarray}
    {\mathbf{\mathcal{E}}}_{\ell+1} &=& {\mathbf{h}}_1^{\ell+1}\ \| \ {\mathbf{h}}_2^{\ell+1}\ \| \ \hdots\ \| \ {\mathbf{h}}_{\mathcal{N}_w + 1}^{\ell+1}.
\end{eqnarray}

\subsection{Training Objective}\label{sec: loss}

It is straight-forward to train a superpoint-referring expression matching network: given ground-truth binary mask of the referring expression $\mathbf{Y} \in \mathbb{R}^{\mathcal{N}_p}$, we first get the corresponding superpoint mask $\mathbf{Y}_s \in \mathbb{R}^{\mathcal{N}_s}$ by superpoint pooling follewed by a 0.5-threshold binarization, and then we apply the binary cross-entropy (BCE) loss on the final response map $\mathbf{\mathcal{M}}$. The operation can be written as:
\begin{align}
    \mathcal{L}_{bce}(\mathbf{\mathcal{M}},\mathbf{Y}_s) &=  BCE(\mathbf{\mathcal{M}},\mathbf{Y}_s),\\
    \mathbf{Y}_s^i &= \mathbb{I}(\sigma(\text{AvgPool}(\mathbf{Y},\mathcal{K}_i))),
\end{align}
where $\text{AvgPool}(\cdot)$ denotes the superpoint average pooling operation, and $\mathbf{Y}_s^i$ denotes the binarized mask value of the $i$-th superpoint $\mathcal{K}_i$. $\mathbb{I}(\cdot)$ indicates whether the mask value is higher than 50\%.

While BCE loss treats each superpoint separately, it falls short in addressing the issue of foreground-background sample imbalance. To tackle this problem, we can use Dice loss~\cite{Milletari_Navab_Ahmadi_2016}:
\begin{equation}
    \mathcal{L}_{dice}(\mathbf{\mathcal{M}},\mathbf{Y}_s) =  1-\frac{2\sum_{i=1}^{\mathcal{N}_s}\mathbf{\mathcal{M}}^i \mathbf{Y}_s^i}{\sum_{i=1}^{\mathcal{N}_s}\mathbf{\mathcal{M}}^i + \sum_{i=1}^{\mathcal{N}_s}\mathbf{Y}_s^i}.
\end{equation}

\begin{table*}[!t]
\footnotesize
\centering
\setlength{\tabcolsep}{4pt}
\setlength{\abovecaptionskip}{2pt}
\resizebox{1\textwidth}{!}{
\begin{tabular}{c|ccc|ccc|ccc|cccc}
\toprule
\rowcolor[HTML]{E6E6E6} &\multicolumn{3}{c|}{Unique ($\sim$19\%)} & \multicolumn{3}{c|}{Multiple ($\sim$81\%)} & \multicolumn{3}{c|}{\textbf{Overall}} & \multicolumn{4}{c}{Inference Time} \\
\rowcolor[HTML]{E6E6E6} \multirow{-2}{*}{Method} & 0.25 & 0.5 & mIoU & 0.25 & 0.5 & mIoU & \textbf{0.25} & \textbf{0.5} & \textbf{mIoU} & Stage-1 & Stage-2 & All & {Speed Up} \\
\midrule
TGNN (GRU) & - & - & - & - & - & - & 35.0   & 29.0  & 26.1 & - & - & - & - \\

TGNN (GRU) \dag & 67.2 & 54.1 & 48.7 & 29.1 & 23.9 & 21.8 & 36.5 & 29.8 & 27.0 & 26139ms & 125ms & 26264ms & 1$\times$\\

3D-STMN(GRU)  & \colorm{88.3} & \colorm{82.8} & \colorm{73.0} & \colorm{45.5} & \colorm{25.8} & \colorm{29.5} & \colorm{53.8} & \colorm{36.8} & \colorm{38.0}  & - & - & 277ms & 97$\times$ \\  
\midrule
TGNN (BERT) & - & - & - & - & - & - & 37.5 & 31.4 & 27.8 & - & - & - & - \\  

TGNN (BERT) \dag & \colorl{69.3} & \colorl{57.8} & \colorl{50.7} & \colorl{31.2} & \colorl{26.6} & \colorl{23.6} & \colorl{38.6} & \colorl{32.7} & \colorl{28.8} & 26862ms & 235ms & 27097ms & 1$\times$ \\   

3D-STMN  & \colorh{\textbf{89.3}} & \colorh{\textbf{84.0}} & \colorh{\textbf{74.5}} & \colorh{\textbf{46.2}} & \colorh{\textbf{29.2}} & \colorh{\textbf{31.1}} & \colorh{\textbf{54.6}}     & \colorh{\textbf{39.8}}  & \colorh{\textbf{39.5}}   & - & - & {\textbf{283ms}} & {\textbf{95.7$\times$}} \\ 
\bottomrule
\end{tabular}}
\caption{The 3D-RES results on ScanRefer, including mIoU and accuracy evaluated by IoU 0.25 and IoU 0.5. \dag~The mIoU and accuracy are reevaluated on our machine. The inference time is measured on the same RTX 1080Ti GPU.}
\label{tab:scanrefer_benchmark}
\end{table*}

\begin{figure*}[!t]
\centering 
  \includegraphics[width=2\columnwidth]{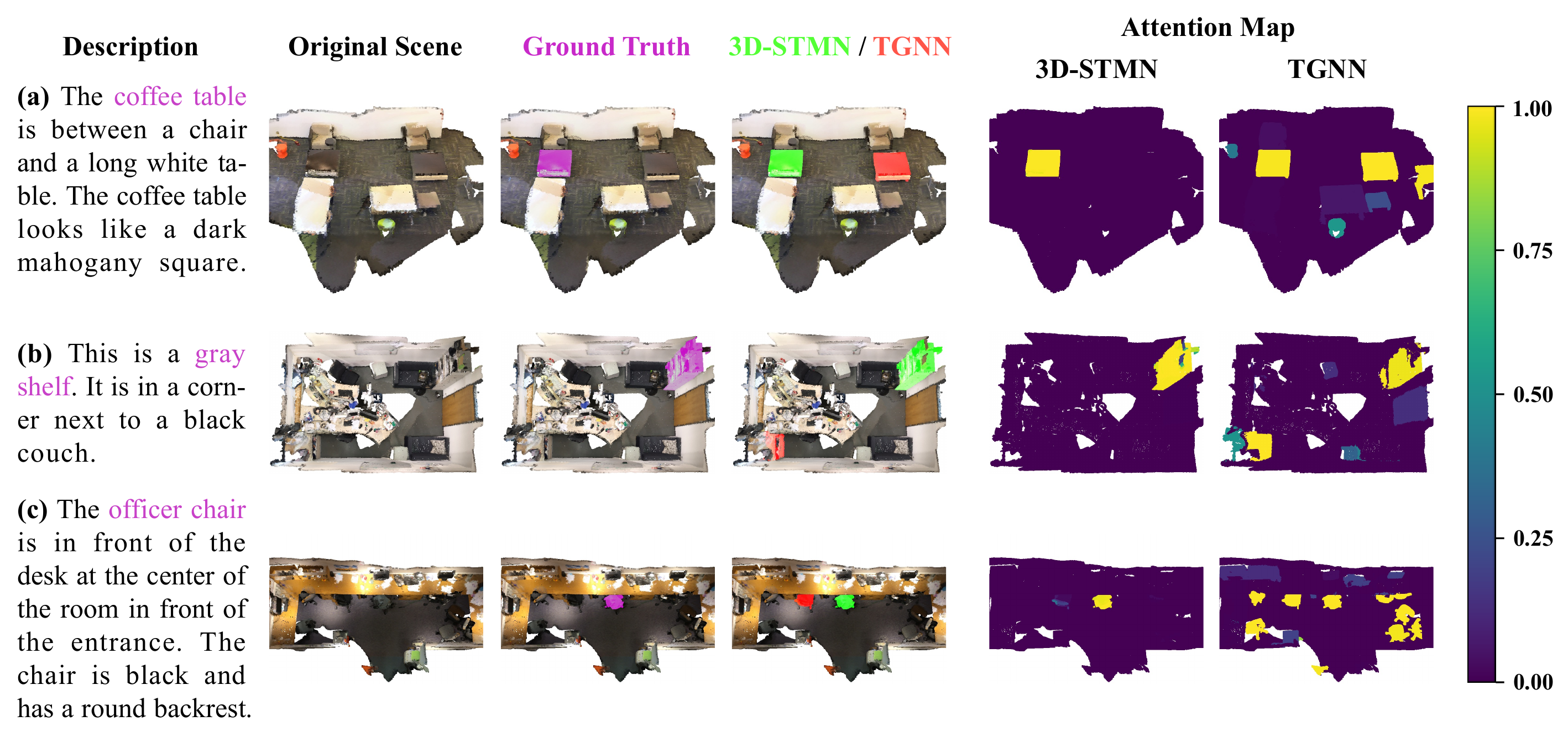}
  \caption{ Visualization of the prediction results and attention maps of our 3D-STMN and TGNN. 
  \textbf{Zoom in for best view.}
  }
  \label{fig: vis}
\end{figure*}

In the STM module, we apply $\mathcal{L}_{rel}$ following~\cite{Luo_Fu_Kong_Gao_Ren_Shen_Xia_Liu_2022} to supervise  the description relevance score $s_r$ with cross-entropy (BCE) loss. The supervision of $s_r$ is based on whether the point belongs to an object category mentioned in the description.

In addition, we add a simple auxiliary score loss $\mathcal{L}_{score}$ for proposal quality prediction following~\cite{Sun_Qing_Tan_Xu_2022}:
\begin{equation}
    \mathcal{L}_{score} =  \mathbb{I}_{\{iou\}} \Vert s-iou \Vert_2,
\end{equation}
where $s$ is the quality score prediction from segmentation kernels. $\mathbb{I}_{\{iou\}}$ indicates whether the IoU between proposal mask prediction and assigned ground truth is higher than 50\%.

Overall, the final training loss function $\mathcal{L}$ can be formulated as:
\begin{equation}
    \mathcal{L} = \lambda_{bce}\mathcal{L}_{bce} + \lambda_{dice}\mathcal{L}_{dice} + \lambda_{rel}\mathcal{L}_{rel} + \lambda_{score}\mathcal{L}_{score},
\end{equation}
where $\lambda_{bce}$, $\lambda_{dice}$, $\lambda_{rel}$ and $\lambda_{score}$ are hyperparameters used to balance these four losses. Empirically, we set $\lambda_{bce}=\lambda_{dice}=1, \lambda_{rel}=5, \lambda_{score}=0.5$.

\section{Experiments}

\subsection{Experiment Settings}

We use the pre-trained Sparse 3D U-Net to extract point-wise features~\cite{Sun_Qing_Tan_Xu_2022}. Meanwhile, we adopt the pre-trained BERT~\cite{Devlin_Chang_Lee_Toutanova_2019} as text encoder following the settings in~\cite{Huang_Lee_Chen_Liu_2021}. The rest of the network is trained from scratch. The initial learning rate is 0.0001. We apply learning rate decay at epoch \{26, 34, 40\} with a rate of 0.5. The number $k_{rel}$ of $\mathbf{\hat{S}}_{rel}$ in STM is set to 512. The default number of multiple rounds $L$ is 6. The batch size is 64, and the maximum sentence length is 80. All experiments are implemented with PyTorch, trained on a single NVIDIA Tesla A100 GPU. 

\subsection{Dataset}

We evaluate our method using the recent 3D referring dataset ScanRefer~\cite{Chen_Chang_Nießner_2020, Huang_Lee_Chen_Liu_2021} which comprises 51,583 natural language expressions that refer to 11,046 objects in 800 ScanNet~\cite{Dai_Chang_Savva_Halber_Funkhouser_Niessner_2017} scenes. 

The evaluation metric is the mean IoU (mIoU) and Acc@$k$IoU, which means the fraction of descriptions whose predicted mask overlaps the ground truth with IoU$>k$, where $k \in \{0.25, 0.5\}$.

\subsection{Quantitative Comparison}
As far as we know, the only existing work that addresses the task of 3D-RES is TGNN~\cite{Huang_Lee_Chen_Liu_2021}. We report the results on the ScanRefer dataset in Tab.~\ref{tab:scanrefer_benchmark}. 

Our proposed 3D-STMN achieves state-of-the-art performance by a substantial margin, with an overall improvement of \textbf{17.1\%}, \textbf{8.4\%}, \textbf{11.7\%} in terms of \textbf{Acc@0.25}, \textbf{Acc@0.5} and \textbf{mIoU}, respectively. In terms of inference speed, we calculated the average inference time for each description. Our 3D-STMN exhibits a significant advantage, being \textbf{95.7} $\times$ faster than the two-stage TGNN. Due to the inference time of our model being controlled within \textbf{0.3 seconds}, it makes \textbf{real-time} applications of 3D-RES possible. 
Our 3D-STMN notably outperforms TGNN, whether using BERT or GRU features, highlighting our model's robustness and inference power. In the ``Unique'' setting, our model boosts Acc@0.25 by 30 points, underscoring its precision with unique objects.

\subsection{Ablation Study}

\subsubsection{STM Mechanism}
We first conducted an ablation study on whether to use superpoints-level features, without the DDI module. As can be seen in Tab.~\ref{tab:stm}, under the same settings, 
the second row outperforms significantly in all metrics, demonstrating the effectiveness of using superpoints as representations.

Next, in rows 3-6, we added the DDI module. Regardless of the structure of the DDI module, it greatly enhances the performance of the segmentation kernels, leading to significant improvements in all metrics, demonstrating fine-grained discriminability of dependency-driven features.

In the STM framework, the choice of segmentation kernel strategy plays a pivotal role in how dependency-driven features construct the kernel for segmentation.
We tested three distinct strategies: \textbf{\romannumeral1)}~Root: This employs the embedding of the root node to formulate the segmentation kernel; \textbf{\romannumeral2)}~Top1: Leverages the word embedding with the highest score, which is derived by averaging the word-superpoint attention map along the superpoint dimension; \textbf{\romannumeral3)}~Average: Utilizes an embedding computed by averaging embeddings of all words. 
Our findings, presented in Tab.~\ref{tab:stm}, reveal that the Top1 strategy emerges as the most effective due to its innate ability to adapt visually. Consequently, we've chosen this setup for subsequent experiments in our study.


\subsubsection{Structure of DDI}

\begin{table}[!t]
\footnotesize
\centering
\setlength{\tabcolsep}{4pt}
\setlength{\abovecaptionskip}{2pt}
\resizebox{1\columnwidth}{!}{
\begin{tabular}{ccc|c|c|ccc}
\toprule
\rowcolor[HTML]{E6E6E6} & & Segmentation &\multicolumn{1}{c|}{Unique} & \multicolumn{1}{c|}{Multiple} & \multicolumn{3}{c}{\textbf{Overall}} \\
\rowcolor[HTML]{E6E6E6} \multirow{-2}{*}{Method} & \multirow{-2}{*}{Superpoint} & Kernel
 & mIoU & mIoU & \textbf{0.25} & \textbf{0.5} & \textbf{mIoU} \\
\midrule
{3D-STMN w/o DDI} & & CLS & 55.3 & 18.5 & 39.4  & 22.0     & 25.6 \\
{3D-STMN w/o DDI} & \ding{51} & CLS & 66.2 & 25.1 & 46.8 & 32.9 & 33.1 \\

\midrule
3D-STMN & \ding{51} &Root & \colorl{72.9} & \colorl{29.6} & \colorm{53.9} & \colorl{37.3} & \colorl{38.0}   \\
3D-STMN & \ding{51} &Avg & \colorm{73.1} & \colorm{30.2} & \colorl{52.2}     & \colorm{39.5} & \colorm{38.6}     \\  
3D-STMN & \ding{51} &Top1 & \colorh{\textbf{74.5}} & \colorh{\textbf{31.1}} & \colorh{\textbf{54.6}}     & \colorh{\textbf{39.8}}  & \colorh{\textbf{39.5}}  \\ 
\bottomrule
\end{tabular}}
\caption{Ablation study of STM, where ``w/o DDI'' denotes directly using the [CLS] token to generate segmentation kernel instead of using the proposed DDI module. }
\label{tab:stm}
\end{table}

\begin{table}[!t]
\footnotesize
\centering
\setlength{\tabcolsep}{4pt}
\setlength{\abovecaptionskip}{2pt}

\scalebox{0.72}{
\begin{tabular}{c|ccc|ccc|ccc}
\toprule
\rowcolor[HTML]{E6E6E6} &\multicolumn{3}{c|}{Unique ($\sim$19\%)} & \multicolumn{3}{c|}{Multiple ($\sim$81\%)} & \multicolumn{3}{c}{\textbf{Overall}} \\
\rowcolor[HTML]{E6E6E6} \multirow{-2}{*}{DDI Structure} & 0.25 & 0.5 & mIoU & 0.25 & 0.5 & mIoU & \textbf{0.25} & \textbf{0.5} & \textbf{mIoU} \\
\midrule
{w/o DDI} & 80.2 & 75.3 & 66.2 & 38.8 & 22.7 & 25.1 & 46.8 & 32.9 & 33.1 \\
\midrule
GA      & \colorm{88.0} & \colorm{83.0} & \colorm{72.8} & 42.0 & 25.5 & 27.6 & \colorl{51.0} & 36.7 & 36.4 \\  
SA~-~GA & 86.9 & 82.3 & 72.6 & \colorm{42.4} & \colorl{27.7} & \colorl{29.1} & \colorm{51.0} & \colorl{38.3} & \colorl{37.5} \\
GA~-~SA & \colorl{87.1} & \colorl{82.6} & \colorl{72.7} & \colorl{42.1} & \colorm{28.4} & \colorm{29.1} & 50.9 & \colorm{38.9} & \colorm{37.6} \\
GA~$\|$~SA & \colorh{\textbf{89.3}} & \colorh{\textbf{84.0}} & \colorh{\textbf{74.5}} & \colorh{\textbf{46.2}} & \colorh{\textbf{29.2}} & \colorh{\textbf{31.1}} & \colorh{\textbf{54.6}}     & \colorh{\textbf{39.8}}  & \colorh{\textbf{39.5}}  \\ 
\bottomrule
\end{tabular}
}
\caption{Ablation study of DDI module, where ``w/o DDI'' denotes not using the proposed DDI module.}
\label{tab:ddi}
\end{table}

In Tab.~\ref{tab:ddi}, we explored four different versions of the structure of DDI module: \textbf{\romannumeral1)} GA (graph-attention only), \textbf{\romannumeral2)} SA~-~GA (series of self-attention followed by graph-attention), \textbf{\romannumeral3)} GA~-~SA (series of graph-attention followed by self-attention), and \textbf{\romannumeral4)} GA~$\|$~SA (graph-attention and self-attention running in parallel).

Our findings reveal that the GA configuration, when compared to the absence of the DDI module, brings about a marked enhancement in performance. This underscores the pivotal role of detailed dependency-driven interactions in our model.
After concatenating SA with dense connections on GA (SA~-~GA and GA~-~SA), the ``Overall'' performance of the model has improved due to the addition of long-range connections, demonstrating the complementary role of SA in enhancing the effectiveness of the GA structure.
Finally, by incorporating parallel self-attention (GA~$\|$~SA), a notable boost in performance was achieved across all settings. This highlights the efficacy of utilizing a parallel connection while simultaneously supplementing long-range connections, which preserves the explicit modeling capability of the dependency tree and maintains the ordered interaction of information.

\subsubsection{Edge Direction of Dependency Graph}
The ablation of edge direction in the dependency graph is presented in Tab.~\ref{tab:direction}. The findings are as follows: 
\textbf{\romannumeral1)} the \textbf{Bi-directional} setting may seem reasonable, but it doubles the number of edge types and leads to overly chaotic information flow, significantly increasing the difficulty of learning.
\textbf{\romannumeral2)} The \textbf{Forward} direction setting outperforms the bidirectional one, but its top-down information flow from the root node leads to inefficient updates in the upper-level nodes, which are often more relevant to the target object.
\textbf{\romannumeral3)} The \textbf{Reverse} setting is optimal because it facilitates bottom-up information flow from leaf nodes to higher-level ones. This results in a progressive accumulation of richer information at each level, mirroring the way humans comprehend complex sentences.

\begin{table}[!t]
\footnotesize
\centering
\setlength{\tabcolsep}{4pt}
\setlength{\abovecaptionskip}{2pt}

\scalebox{0.72}{
\begin{tabular}{c|ccc|ccc|ccc}
\toprule
\rowcolor[HTML]{E6E6E6} &\multicolumn{3}{c|}{Unique ($\sim$19\%)} & \multicolumn{3}{c|}{Multiple ($\sim$81\%)} & \multicolumn{3}{c}{\textbf{Overall}} \\
\rowcolor[HTML]{E6E6E6} \multirow{-2}{*}{Edge Direction} & 0.25 & 0.5 & mIoU & 0.25 & 0.5 & mIoU & \textbf{0.25} & \textbf{0.5} & \textbf{mIoU} \\
\midrule
Bi-directional & \colorl{87.4} & \colorl{82.3} & \colorl{72.7} & \colorl{42.6} & \colorl{28.2} & \colorl{29.4} & \colorl{51.3} & \colorl{38.7} & \colorl{37.8} \\
Forward      & \colorm{89.1} & \colorm{83.9} & \colorm{74.2} & \colorm{45.8} & \colorm{28.6} & \colorm{30.8} & \colorm{54.2} & \colorm{39.3} & \colorm{39.2} \\  
Reverse & \colorh{\textbf{89.3}} & \colorh{\textbf{84.0}} & \colorh{\textbf{74.5}} & \colorh{\textbf{46.2}} & \colorh{\textbf{29.2}} & \colorh{\textbf{31.1}} & \colorh{\textbf{54.6}}     & \colorh{\textbf{39.8}}  & \colorh{\textbf{39.5}}  \\ 
\bottomrule
\end{tabular}
}
\caption{Analyzing the edge direction of Dependency Graph.}
\vspace{-0.3cm}
\label{tab:direction}
\end{table}

\begin{table}[!t]
\footnotesize
\centering
\setlength{\tabcolsep}{4pt}
\setlength{\abovecaptionskip}{2pt}

\scalebox{0.72}{
\begin{tabular}{c|ccc|ccc|ccc}
\toprule
\rowcolor[HTML]{E6E6E6} Sampling &\multicolumn{3}{c|}{Unique ($\sim$19\%)} & \multicolumn{3}{c|}{Multiple ($\sim$81\%)} & \multicolumn{3}{c}{\textbf{Overall}} \\
\rowcolor[HTML]{E6E6E6}  Number & 0.25 & 0.5 & mIoU & 0.25 & 0.5 & mIoU & \textbf{0.25} & \textbf{0.5} & \textbf{mIoU} \\
\midrule
64 & 86.5 & 80.8 & 71.7 & 42.3 & 24.9 & 28.0 & 50.9 & 35.7 & 36.5 \\
128 & \colorl{87.3} & \colorm{81.7} & \colorl{72.5} & \colorl{43.1} & 25.3 & 27.9 & \colorl{51.7} & 36.2 & 36.6 \\  
256 & 85.6 & 81.2 & 71.9 & 42.2 & \colorl{28.0} & 29.0 & 50.6 & \colorl{38.3} & 37.3 \\
512 & \colorh{\textbf{89.3}} & \colorh{\textbf{84.0}} & \colorh{\textbf{74.5}} & \colorh{\textbf{46.2}} & \colorh{\textbf{29.2}} & \colorh{\textbf{31.1}} & \colorh{\textbf{54.6}}     & \colorh{\textbf{39.8}}  & \colorh{\textbf{39.5}}  \\ 
1024 & 87.2 & 81.5 & 72.2 & \colorm{43.6} & 27.3 & \colorm{29.4} & \colorm{52.0} & 37.8 & \colorm{37.8} \\
\midrule
w/o sampling & \colorm{87.4} & \colorl{81.6} & \colorm{72.6} & 42.0 & \colorm{28.7} & \colorl{29.4} & 50.8 & \colorm{39.0} & \colorl{37.8} \\
\bottomrule
\end{tabular}
}
\caption{Ablation study of sampling number of superpoints, where ``w/o sampling'' means using all superpoints.}
\label{tab:sample_number}
\vspace{-1em}
\end{table}

\subsubsection{Sampling Number of Superpoints}
Within the ScanRefer dataset, the count of superpoints highlighted in the description varies. Investigating the optimal sampling number in STM is vital. As displayed in Tab.~\ref{tab:sample_number}, our model's performance initially rises with increased superpoints, peaking at 512, then declines. Notably, using our sampling strategy yields significantly better results than not sampling at all.

\subsection{Qualitative Comparison}
In this subsection, we perform a qualitative comparison on ScanRefer validation set, demonstrating the remarkable discriminative ability of our 3D-STMN compared to TGNN. Fig.~\ref{fig: vis} visually demonstrates the superior performance of our 3D-STMN in accurately localizing target objects on attention maps, regardless of the difficulty level of the test samples. The attention generated by 3D-STMN is highly focused and exhibits remarkable precision. Conversely, TGNN struggles with discernment, as it exhibits significantly high attention values for multiple semantically similar objects, as observed in cases \textbf{(a)}, \textbf{(b)} and \textbf{(c)}. Notably, when confronted with scenes containing multiple objects similar to the target, accompanied by longer and more complex textual descriptions as seen in cases \textbf{(a)} and \textbf{(c)}, TGNN fails to distinguish and accurately localize the target, rendering its performance comparable to random guessing. In contrast, our 3D-STMN is capable of precise segmentation for these challenging samples. Similar to humans, it exhibits a subtle but distinct focus on objects closely adjacent to the target, distinguishing them from the background, as in case \textbf{(c)}.
\section{Conclusion}
We present 3D-STMN, an efficient and dense-aligned end-to-end method for 3D-RES. 
By employing the Superpoint-Text Matching (STM) mechanism, our model successfully breaks free from the limitations of the traditional two-stage paradigm. This liberates us to leverage end-to-end dense supervision, harnessing the advantages of precise segmentation and rapid inference speed. Specifically, our model achieves an impressive inference speed of less than 1 second per scene, rendering it well-suited for real-time applications and highly applicable in time-critical scenarios. Furthermore, the proposed Dependency-Driven Interaction (DDI) module substantially enhances our model's comprehension of referring expressions. By explicitly modeling dependency relationships, our model exhibits improved localization and segmentation capabilities, demonstrating a significant advancement in performance. 

{\small
\bibliographystyle{ieee_fullname}
\bibliography{egbib}

\begin{thebibliography}{10}\itemsep=-1pt

\bibitem{Achanta_Shaji_Smith_Lucchi_Fua_Süsstrunk_2012}
Radhakrishna Achanta, Appu Shaji, Kevin Smith, Aurelien Lucchi, Pascal Fua, and
  Sabine S{\"u}sstrunk.
\newblock Slic superpixels compared to state-of-the-art superpixel methods.
\newblock {\em IEEE transactions on pattern analysis and machine intelligence},
  34(11):2274--2282, 2012.

\bibitem{Achlioptas_Abdelreheem_Xia_Elhoseiny_Guibas_2020}
Panos Achlioptas, Ahmed Abdelreheem, Fei Xia, Mohamed Elhoseiny, and Leonidas
  Guibas.
\newblock Referit3d: Neural listeners for fine-grained 3d object identification
  in real-world scenes.
\newblock In {\em Computer Vision--ECCV 2020: 16th European Conference,
  Glasgow, UK, August 23--28, 2020, Proceedings, Part I 16}, pages 422--440.
  Springer, 2020.

\bibitem{Chen_Chang_Nießner_2020}
Dave~Zhenyu Chen, Angel~X Chang, and Matthias Nie{\ss}ner.
\newblock Scanrefer: 3d object localization in rgb-d scans using natural
  language.
\newblock In {\em European conference on computer vision}, pages 202--221.
  Springer, 2020.

\bibitem{Dai_Chang_Savva_Halber_Funkhouser_Niessner_2017}
Angela Dai, Angel~X Chang, Manolis Savva, Maciej Halber, Thomas Funkhouser, and
  Matthias Nie{\ss}ner.
\newblock Scannet: Richly-annotated 3d reconstructions of indoor scenes.
\newblock In {\em Proceedings of the IEEE conference on computer vision and
  pattern recognition}, pages 5828--5839, 2017.

\bibitem{Deng_Wu_Wu_Hu_Lyu_Tan_2018}
Chaorui Deng, Qi Wu, Qingyao Wu, Fuyuan Hu, Fan Lyu, and Mingkui Tan.
\newblock Visual grounding via accumulated attention.
\newblock In {\em Proceedings of the IEEE conference on computer vision and
  pattern recognition}, pages 7746--7755, 2018.

\bibitem{Devlin_Chang_Lee_Toutanova_2019}
Jacob Devlin, Ming-Wei Chang, Kenton Lee, and Kristina Toutanova.
\newblock Bert: Pre-training of deep bidirectional transformers for language
  understanding.
\newblock {\em arXiv preprint arXiv:1810.04805}, 2018.

\bibitem{Ding_Liu_Wang_Jiang_2021}
Henghui Ding, Chang Liu, Suchen Wang, and Xudong Jiang.
\newblock Vision-language transformer and query generation for referring
  segmentation.
\newblock In {\em Proceedings of the IEEE/CVF International Conference on
  Computer Vision}, pages 16321--16330, 2021.

\bibitem{Ding_Ding_Hui_Huang_Wei_Wei_Liu_2022}
Zihan Ding, Zi-han Ding, Tianrui Hui, Junshi Huang, Xiaoming Wei, Xiaolin Wei,
  and Si Liu.
\newblock Ppmn: Pixel-phrase matching network for one-stage panoptic narrative
  grounding.
\newblock In {\em Proceedings of the 30th ACM International Conference on
  Multimedia}, pages 5537--5546, 2022.

\bibitem{Dwivedi_Bresson_2020}
Vijay~Prakash Dwivedi and Xavier Bresson.
\newblock A generalization of transformer networks to graphs.
\newblock {\em arXiv preprint arXiv:2012.09699}, 2020.

\bibitem{Engelmann_Bokeloh_Fathi_Leibe_NieBner_2020}
Francis Engelmann, Martin Bokeloh, Alireza Fathi, Bastian Leibe, and Matthias
  Nie{\ss}ner.
\newblock 3d-mpa: Multi-proposal aggregation for 3d semantic instance
  segmentation.
\newblock In {\em Proceedings of the IEEE/CVF conference on computer vision and
  pattern recognition}, pages 9031--9040, 2020.

\bibitem{Feng_Li_Li_Zhang_Zhang_Zhu_Zhang_Wang_Mian_2021}
Mingtao Feng, Zhen Li, Qi Li, Liang Zhang, XiangDong Zhang, Guangming Zhu, Hui
  Zhang, Yaonan Wang, and Ajmal Mian.
\newblock Free-form description guided 3d visual graph network for object
  grounding in point cloud.
\newblock In {\em Proceedings of the IEEE/CVF International Conference on
  Computer Vision}, pages 3722--3731, 2021.

\bibitem{Graham_Engelcke_Maaten_2018}
Benjamin Graham, Martin Engelcke, and Laurens Van Der~Maaten.
\newblock 3d semantic segmentation with submanifold sparse convolutional
  networks.
\newblock In {\em Proceedings of the IEEE conference on computer vision and
  pattern recognition}, pages 9224--9232, 2018.

\bibitem{Han_Zheng_Xu_Fang_2020}
Lei Han, Tian Zheng, Lan Xu, and Lu Fang.
\newblock Occuseg: Occupancy-aware 3d instance segmentation.
\newblock In {\em Proceedings of the IEEE/CVF conference on computer vision and
  pattern recognition}, pages 2940--2949, 2020.

\bibitem{He_Li_Li_Zhang}
Chenhang He, Ruihuang Li, Shuai Li, and Lei Zhang.
\newblock Voxel set transformer: A set-to-set approach to 3d object detection
  from point clouds.
\newblock In {\em Proceedings of the IEEE/CVF Conference on Computer Vision and
  Pattern Recognition}, pages 8417--8427, 2022.

\bibitem{Hu_Rohrbach_Andreas_Darrell_Saenko_2017}
Ronghang Hu, Marcus Rohrbach, Jacob Andreas, Trevor Darrell, and Kate Saenko.
\newblock Modeling relationships in referential expressions with compositional
  modular networks.
\newblock In {\em Proceedings of the IEEE conference on computer vision and
  pattern recognition}, pages 1115--1124, 2017.

\bibitem{Hu_Rohrbach_Darrell_2016}
Ronghang Hu, Marcus Rohrbach, and Trevor Darrell.
\newblock Segmentation from natural language expressions.
\newblock In {\em Computer Vision--ECCV 2016: 14th European Conference,
  Amsterdam, The Netherlands, October 11--14, 2016, Proceedings, Part I 14},
  pages 108--124. Springer, 2016.

\bibitem{Huang_Lee_Chen_Liu_2021}
Pin-Hao Huang, Han-Hung Lee, Hwann-Tzong Chen, and Tyng-Luh Liu.
\newblock Text-guided graph neural networks for referring 3d instance
  segmentation.
\newblock In {\em Proceedings of the AAAI Conference on Artificial
  Intelligence}, volume~35, pages 1610--1618, 2021.

\bibitem{Jain_Gkanatsios_Mediratta_Fragkiadaki_2021}
Ayush Jain, Nikolaos Gkanatsios, Ishita Mediratta, and Katerina Fragkiadaki.
\newblock Bottom up top down detection transformers for language grounding in
  images and point clouds.
\newblock In {\em European Conference on Computer Vision}, pages 417--433.
  Springer, 2022.

\bibitem{Landrieu_Simonovsky_2018}
Loic Landrieu and Martin Simonovsky.
\newblock Large-scale point cloud semantic segmentation with superpoint graphs.
\newblock In {\em Proceedings of the IEEE conference on computer vision and
  pattern recognition}, pages 4558--4567, 2018.

\bibitem{Li_Li_Kuo_Shu_Qi_Shen_Jia_2018}
Ruiyu Li, Kaican Li, Yi-Chun Kuo, Michelle Shu, Xiaojuan Qi, Xiaoyong Shen, and
  Jiaya Jia.
\newblock Referring image segmentation via recurrent refinement networks.
\newblock In {\em Proceedings of the IEEE Conference on Computer Vision and
  Pattern Recognition}, pages 5745--5753, 2018.

\bibitem{Liang_Li_Xu_Tan_Jia_2021}
Zhihao Liang, Zhihao Li, Songcen Xu, Mingkui Tan, and Kui Jia.
\newblock Instance segmentation in 3d scenes using semantic superpoint tree
  networks.
\newblock In {\em Proceedings of the IEEE/CVF International Conference on
  Computer Vision}, pages 2783--2792, 2021.

\bibitem{Lin_Wang_Zhai_Li_Li_2018}
Yangbin Lin, Cheng Wang, Dawei Zhai, Wei Li, and Jonathan Li.
\newblock Toward better boundary preserved supervoxel segmentation for 3d point
  clouds.
\newblock {\em ISPRS journal of photogrammetry and remote sensing}, 143:39--47,
  2018.

\bibitem{Liu_Zhang_Zha_Wu_2019}
Daqing Liu, Hanwang Zhang, Feng Wu, and Zheng-Jun Zha.
\newblock Learning to assemble neural module tree networks for visual
  grounding.
\newblock In {\em Proceedings of the IEEE/CVF International Conference on
  Computer Vision}, pages 4673--4682, 2019.

\bibitem{luo2020cascade}
Gen Luo, Yiyi Zhou, Rongrong Ji, Xiaoshuai Sun, Jinsong Su, Chia-Wen Lin, and
  Qi Tian.
\newblock Cascade grouped attention network for referring expression
  segmentation.
\newblock In {\em Proceedings of the 28th ACM International Conference on
  Multimedia}, pages 1274--1282, 2020.

\bibitem{luo2020multi}
Gen Luo, Yiyi Zhou, Xiaoshuai Sun, Liujuan Cao, Chenglin Wu, Cheng Deng, and
  Rongrong Ji.
\newblock Multi-task collaborative network for joint referring expression
  comprehension and segmentation.
\newblock In {\em Proceedings of the IEEE/CVF Conference on computer vision and
  pattern recognition}, pages 10034--10043, 2020.

\bibitem{Luo_Fu_Kong_Gao_Ren_Shen_Xia_Liu_2022}
Junyu Luo, Jiahui Fu, Xianghao Kong, Chen Gao, Haibing Ren, Hao Shen, Huaxia
  Xia, and Si Liu.
\newblock 3d-sps: Single-stage 3d visual grounding via referred point
  progressive selection.
\newblock In {\em Proceedings of the IEEE/CVF Conference on Computer Vision and
  Pattern Recognition}, pages 16454--16463, 2022.

\bibitem{Manning_Surdeanu_Bauer_Finkel_Bethard_McClosky_2014}
Christopher~D Manning, Mihai Surdeanu, John Bauer, Jenny~Rose Finkel, Steven
  Bethard, and David McClosky.
\newblock The stanford corenlp natural language processing toolkit.
\newblock In {\em Proceedings of 52nd annual meeting of the association for
  computational linguistics: system demonstrations}, pages 55--60, 2014.

\bibitem{Margffoy-Tuay_Pérez_Botero_Arbeláez_2018}
Edgar Margffoy-Tuay, Juan~C P{\'e}rez, Emilio Botero, and Pablo Arbel{\'a}ez.
\newblock Dynamic multimodal instance segmentation guided by natural language
  queries.
\newblock In {\em Proceedings of the European Conference on Computer Vision
  (ECCV)}, pages 630--645, 2018.

\bibitem{Milletari_Navab_Ahmadi_2016}
Fausto Milletari, Nassir Navab, and Seyed-Ahmad Ahmadi.
\newblock V-net: Fully convolutional neural networks for volumetric medical
  image segmentation.
\newblock In {\em 2016 fourth international conference on 3D vision (3DV)},
  pages 565--571. Ieee, 2016.

\bibitem{Nagaraja_Morariu_Davis_2016}
Varun~K Nagaraja, Vlad~I Morariu, and Larry~S Davis.
\newblock Modeling context between objects for referring expression
  understanding.
\newblock In {\em Computer Vision--ECCV 2016: 14th European Conference,
  Amsterdam, The Netherlands, October 11--14, 2016, Proceedings, Part IV 14},
  pages 792--807. Springer, 2016.

\bibitem{Papon_Abramov_Schoeler_Worgotter_2013}
Jeremie Papon, Alexey Abramov, Markus Schoeler, and Florentin Worgotter.
\newblock Voxel cloud connectivity segmentation-supervoxels for point clouds.
\newblock In {\em Proceedings of the IEEE conference on computer vision and
  pattern recognition}, pages 2027--2034, 2013.

\bibitem{Robert_Raguet_Landrieu_2023}
Damien Robert, Hugo Raguet, and Loic Landrieu.
\newblock Efficient 3d semantic segmentation with superpoint transformer.
\newblock {\em arXiv preprint arXiv:2306.08045}, 2023.

\bibitem{Sadhu_Chen_Nevatia_2019}
Arka Sadhu, Kan Chen, and Ram Nevatia.
\newblock Zero-shot grounding of objects from natural language queries.
\newblock In {\em Proceedings of the IEEE/CVF International Conference on
  Computer Vision}, pages 4694--4703, 2019.

\bibitem{Shi_Li_Meng_Wu_2018}
Hengcan Shi, Hongliang Li, Fanman Meng, and Qingbo Wu.
\newblock Key-word-aware network for referring expression image segmentation.
\newblock In {\em Proceedings of the European Conference on Computer Vision
  (ECCV)}, pages 38--54, 2018.

\bibitem{Sun_Qing_Tan_Xu_2022}
Jiahao Sun, Chunmei Qing, Junpeng Tan, and Xiangmin Xu.
\newblock Superpoint transformer for 3d scene instance segmentation.
\newblock In {\em Proceedings of the AAAI Conference on Artificial
  Intelligence}, volume~37, pages 2393--2401, 2023.

\bibitem{Tu_Liu_Jampani_Sun_Chien_Yang_Kautz_2018}
Wei-Chih Tu, Ming-Yu Liu, Varun Jampani, Deqing Sun, Shao-Yi Chien, Ming-Hsuan
  Yang, and Jan Kautz.
\newblock Learning superpixels with segmentation-aware affinity loss.
\newblock In {\em Proceedings of the IEEE Conference on Computer Vision and
  Pattern Recognition}, pages 568--576, 2018.

\bibitem{Wang_Ye_Cao_Huang_Sun_He_Tao}
Yikai Wang, TengQi Ye, Lele Cao, Wenbing Huang, Fuchun Sun, Fengxiang He, and
  Dacheng Tao.
\newblock Bridged transformer for vision and point cloud 3d object detection.
\newblock In {\em Proceedings of the IEEE/CVF Conference on Computer Vision and
  Pattern Recognition}, pages 12114--12123, 2022.

\bibitem{Wu_Cheng_Zhang_Cheng_Zhang_2022}
Yanmin Wu, Xinhua Cheng, Renrui Zhang, Zesen Cheng, and Jian Zhang.
\newblock Eda: Explicit text-decoupling and dense alignment for 3d visual
  grounding.
\newblock In {\em Proceedings of the IEEE/CVF Conference on Computer Vision and
  Pattern Recognition}, pages 19231--19242, 2023.

\bibitem{Yang_Li_Yu_2020}
Sibei Yang, Guanbin Li, and Yizhou Yu.
\newblock Graph-structured referring expression reasoning in the wild.
\newblock In {\em Proceedings of the IEEE/CVF conference on computer vision and
  pattern recognition}, pages 9952--9961, 2020.

\bibitem{yang2022lavt}
Zhao Yang, Jiaqi Wang, Yansong Tang, Kai Chen, Hengshuang Zhao, and Philip~HS
  Torr.
\newblock Lavt: Language-aware vision transformer for referring image
  segmentation.
\newblock In {\em Proceedings of the IEEE/CVF Conference on Computer Vision and
  Pattern Recognition}, pages 18155--18165, 2022.

\bibitem{Yang_Zhang_Wang_Luo_2021}
Zhengyuan Yang, Songyang Zhang, Liwei Wang, and Jiebo Luo.
\newblock Sat: 2d semantics assisted training for 3d visual grounding.
\newblock In {\em Proceedings of the IEEE/CVF International Conference on
  Computer Vision}, pages 1856--1866, 2021.

\bibitem{Ye_Rochan_Liu_Wang_2019}
Linwei Ye, Mrigank Rochan, Zhi Liu, and Yang Wang.
\newblock Cross-modal self-attention network for referring image segmentation.
\newblock In {\em Proceedings of the IEEE/CVF conference on computer vision and
  pattern recognition}, pages 10502--10511, 2019.

\bibitem{Yu_Lin_Shen_Yang_Lu_Bansal_Berg_2018}
Licheng Yu, Zhe Lin, Xiaohui Shen, Jimei Yang, Xin Lu, Mohit Bansal, and
  Tamara~L Berg.
\newblock Mattnet: Modular attention network for referring expression
  comprehension.
\newblock In {\em Proceedings of the IEEE conference on computer vision and
  pattern recognition}, pages 1307--1315, 2018.

\bibitem{Yu_Poirson_Yang_Berg_Berg_2016}
Licheng Yu, Patrick Poirson, Shan Yang, Alexander~C Berg, and Tamara~L Berg.
\newblock Modeling context in referring expressions.
\newblock In {\em Computer Vision--ECCV 2016: 14th European Conference,
  Amsterdam, The Netherlands, October 11-14, 2016, Proceedings, Part II 14},
  pages 69--85. Springer, 2016.

\bibitem{Yu_Tan_Bansal_Berg_2017}
Licheng Yu, Hao Tan, Mohit Bansal, and Tamara~L Berg.
\newblock A joint speaker-listener-reinforcer model for referring expressions.
\newblock In {\em Proceedings of the IEEE conference on computer vision and
  pattern recognition}, pages 7282--7290, 2017.

\bibitem{Yuan_Yan_Liao_Zhang_Wang_Li_Cui_2021}
Zhihao Yuan, Xu Yan, Yinghong Liao, Ruimao Zhang, Sheng Wang, Zhen Li, and
  Shuguang Cui.
\newblock Instancerefer: Cooperative holistic understanding for visual
  grounding on point clouds through instance multi-level contextual referring.
\newblock In {\em Proceedings of the IEEE/CVF International Conference on
  Computer Vision}, pages 1791--1800, 2021.

\bibitem{Zhao_Cai_Sheng_Xu_2021}
Lichen Zhao, Daigang Cai, Lu Sheng, and Dong Xu.
\newblock 3dvg-transformer: Relation modeling for visual grounding on point
  clouds.
\newblock In {\em Proceedings of the IEEE/CVF International Conference on
  Computer Vision}, pages 2928--2937, 2021.

\bibitem{Zhuang_Wu_Shen_Reid_Hengel_2017}
Bohan Zhuang, Qi Wu, Chunhua Shen, Ian Reid, and Anton Van Den~Hengel.
\newblock Parallel attention: A unified framework for visual object discovery
  through dialogs and queries.
\newblock In {\em Proceedings of the IEEE Conference on Computer Vision and
  Pattern Recognition}, pages 4252--4261, 2018.

\end{thebibliography}
}

\end{document}